\documentclass[conference,letterpaper]{IEEEtran}
\usepackage{fancyhdr}
\usepackage{graphicx}
\usepackage{amsmath}
\usepackage{amssymb}
\usepackage{algorithmic}
\usepackage{algorithm}
\usepackage{subfigure}
\usepackage{pdfsync}
\usepackage{fancyhdr}

\newcommand{\bfx}{{\textbf{x}}}
\newcommand{\bfv}{{\textbf{v}}}

\newcommand{\bfw}{{\textbf{w}}}
\newcommand{\bfy}{{\textbf{y}}}

\newcommand{\bfalpha}{{\boldsymbol{\alpha}}}
\newcommand{\bftau}{{\boldsymbol{\tau}}}
\newcommand{\bfdelta}{{\boldsymbol{\delta}}}

\begin{document}
\title{Multiple kernel multivariate performance learning using cutting plane algorithm}

\author{
\IEEEauthorblockN{Jingbin Wang}
\IEEEauthorblockA{
National Time Service Center, \\
Chinese Academy of Sciences,\\
Xi' an 710600 , China\\
Graduate University of\\
Chinese Academy of Sciences,\\
 Beijing 100039, China\\
jingbinwang1@outlook.com
}
\and
\IEEEauthorblockN{Haoxiang Wang}
\IEEEauthorblockA{Department of Electrical\\
and Computer Engineering,\\
Cornell University,\\
Ithaca, NY 14850, USA}
\and
\IEEEauthorblockN{Yihua Zhou}
\IEEEauthorblockA{Department of Mechanical\\
Engineering and Mechanics,\\
Lehigh University,\\
Bethlehem, PA 18015, USA}
\and
\IEEEauthorblockN{Nancy McDonald}
\IEEEauthorblockA{
Department of Computer\\
Science, Tulane University,\\
New Orleans, LA 70118, USA\\
nancya.mcdonald@yahoo.com}
}

\IEEEoverridecommandlockouts
\IEEEpubid{\makebox[\columnwidth]{978-1-4799-0652-9/13/\$31.00~\copyright2013
IEEE \hfill} \hspace{\columnsep}\makebox[\columnwidth]{ }}

\maketitle
\thispagestyle{plain}

\fancypagestyle{plain}{
\fancyhf{}	% clear all header and footer fields
\fancyfoot[L]{}
\fancyfoot[C]{}
\fancyfoot[R]{}
\renewcommand{\headrulewidth}{0pt}
\renewcommand{\footrulewidth}{0pt}
}

\pagestyle{fancy}{
\fancyhf{}
\fancyfoot[R]{}}
\renewcommand{\headrulewidth}{0pt}
\renewcommand{\footrulewidth}{0pt}

\begin{abstract}
In this paper, we propose a  multi-kernel classifier learning algorithm to optimize a given nonlinear and nonsmoonth multivariate classifier performance measure.
Moreover, to solve the problem of kernel function selection and kernel parameter tuning, we proposed to construct an optimal kernel by weighted linear combination of some candidate kernels. The learning of the classifier parameter and the kernel weight are unified in a single objective function considering to minimize the upper boundary of the given multivariate performance measure. The objective function is optimized with regard to classifier parameter and kernel weight alternately in an iterative algorithm by using cutting plane algorithm. The developed algorithm is evaluated on two different pattern classification methods with regard to various multivariate performance measure optimization problems. The experiment results show the proposed algorithm outperforms the competing methods.
\end{abstract}


\IEEEpeerreviewmaketitle

\section{Introduction}

In different pattern classification problems, various performances are employed to evaluate the classifiers, including classification accuracy (ACC), F1 score , Matthews Correlation Coefficient (MCC), area under the receiver operating characteristic (ROC) Curve (AUC) and recall-precision break even point (RP-BEP) of recall-precision curve.
Due to the nonlinear and nonsmooth nature of many performance measures, it is difficult to optimize them directly to learn an optimal classifier.  To solve this problem, Joachims \cite{Joachims2005377} proposed a support vector machine learning method  for multivariate Performance measures (SVM$^{Perf}$).
This other method has been applied to optimized some nonlinear multivariate performance measures to learn linear classifiers successfully. However, it is limited to the learning of linear classifiers.
When data samples of different classes cannot be separated by a linear boundary, it is suggested to employ the kernel trick to map the data samples to a nonlinear high-dimensional data space so that a linear boundary could be learned \cite{Mller2001181,wang2009trajectory,Wang201585}. Joachims and Yu \cite{Joachims2009179} also extended the SVM$^{Perf}$ to its kernel version to handle the nonlinearly distributed data.
One important shortage of this method lies on the choosing of an optimal kernel function with its corresponding parameter. In \cite{Joachims2009179}, the RBF-Kernel is used to classification problems on some data sets without any justification, but it is highly doubt if this kernel is suitable for other data sets. Moreover, how the optimal parameter of the kernel function possibly influences the results significantly. One possible way to solve this problem is to conduct an exhausting linear search or a cross validation in the kernel function and parameter space by using the training set, which is very time-consuming and also makes the learned classifier over-fitting to the training samples.

To solve this problem, we assume that the desired kernel can be obtained by the linear combination of some candidate kernel functions with different kernel parameters. The optimal kernel is parameterized by the linear combination weights associated with different kernels. This framework is called Multi-Kernel Learning (MKL) since we explore the nonlinear kernel spaces of multiple kernels \cite{wang2014feature}. To learn the kernel weights, we cast the MKL problem with the multivariate performance measures problem, and proposed an unified learning problem for both MKL and multivariate performance measures problems.
For the first time, we propose the problem of learning an optimal kernel for multivariate performance measures, and a novel solution for this problem by learning kernel in multiple kernel spaces simultaneously  with optimizing multivariate performance measures.

\IEEEpubidadjcol

The rest parts of this paper are organized as follows: in section \ref{sec:method}, we introduce the novel method by formulating the problem first, optimizing it then, and developing an iterative algorithm finally, in section \ref{sec:exp}, the proposed method is evaluated on some benchmark data sets, and in section \ref{sec:con}, the paper is concluded.

\section{Proposed method}
\label{sec:method}

\subsection{Problem Formulation}

We assume we have a training data set with $n$ training samples,
and the training samples are organized in an training matrix
$X=[\bfx_1,\cdots,\bfx_n]\in \mathbb{R}^{d\times n}$,
where the $i$-th column  $\bfx_i$ is the $d$-dimensional feature vector
of the $i$-th training sample.
Moreover, we also organize the class labels in a class label vector
$\bfy=[y_1,\cdots,y_n]^\top\in \{+1,-1\}^{n}$,
where $y_i\in \{+1,-1\}$ is the binary class label of the $i$-th training sample.
Under the framework of kernel learning \cite{wang2014effective}, an sample vector can be mapped into a high dimensional nonlinear Hilbert Space, via a implicit mapping function
$\phi: \bfx\rightarrow \phi(\bfx) \in \mathcal{R}^{d'}$,
where $d'\gg d$ is the dimension of the Hilbert Space.
The mapping function is explored by a kernel function, which is defined as the
dot-produce of the mapping of two samples $\bfx_i$
and $\bfx_j$, as
$K(\bfx_i,\bfx_j) = \phi(\bfx_i)^\top \phi(\bfx_j)$. In the multi-kernel learning framework, we may have several such Hilbert Spaces available and there corresponding nonlinear mapping functions are denoted as
$\{\phi_m(\bfx) \in \mathcal{R}^{d'_m}\}_{m=1}^M$, where $M$ is the number of Hilbert Spaces, $\phi_m(\bfx) $ is the nonlinear mapping function of the $m$-th mapping function, and $d'_m$ is the dimension of the $m$-th Hilbert Space.
We also define the kernel function for the $m$-th Hilbert space as $K_m(\bfx_i,\bfx_j) = \phi(\bfx_i)_m^\top \phi_m(\bfx_j)$. We weight and concatenate the mapping function to form a longer vector in a more general
Hilbert Space,
$\phi_\bftau(\bfx) =
\left [
\tau_1\phi_1(\bfx)^\top, \cdots, \tau_M\phi_M(\bfx)^\top
\right ]^\top
\in \mathbb{R}^{d'}
$
where $\tau_m \in \mathbb{R}_+$ is the nonnegative weight for the $m$-th Hilbert Space,
$\bftau=[\tau_1,\cdots,\tau_M]^\top \in \mathbb{R}_+^M$
is the weight vector,
and $d'=\sum_{m=1}^M d'_m$
is the dimension of the general Hilbert Space.
Its corresponding kernel function is given as

\begin{equation}
\label{equ:Ker1}
\begin{aligned}
K_\bftau(\bfx_i,\bfx_j) = \phi_\bftau(\bfx_i)^\top \phi_\bftau(\bfx_j)=
\sum_{m=1}^M  \tau_m^2 K_m(\bfx_i,\bfv_j)
\end{aligned}
\end{equation}
It can be seen that the kernel function is also a weighted linear combination of the
$M$ kernel functions of the $M$ Hilbert spaces.
We map all the samples to the Hilbert spaces,
and organize the mapping results in a $d' \times n$ matrix as
$\phi_\bftau(X)=[\phi_\bftau(\bfx_1),\cdots,\phi_\bftau(\bfx_n)]\in \mathbb{R}^{d'\times n}$.
We can also apply the kernel function to the matrix and obtain
the $n\times n$ kernel matrix
$K_\bftau(X,X)=\sum_{m=1}^M  \tau_m^2 K_m(X,X)\in \mathbb{R}^{n\times n}$,
where
$K_m(X,X)=[K_m(\bfx_i,\bfx_j)]\in \mathbb{R}^{n\times n}$
is the kernel matrix of the $m$-th Hilbert space.

We consider the problem of learning a hypotheses function  $h_\bfw(X)$ which maps a tuple of $n$ samples organized in a data matrix $X$ to a label vector of $n$ labels $\bfy$.
To this end, we first map the data matrix $X$ to the general Hilbert space
$\phi_\bftau(X)$,
and then apply a
linear discriminant
function of the following form

\begin{equation}
\label{equ:h_w}
\begin{aligned}
h_\bfw(X)
&=\underset{\bfy'\in  \{+1,-1\}^{n}} {\arg\max}\bfw^\top \phi_\bftau(X) \bfy'=\underset{\bfy'\in  \{+1,-1\}^{n}} {\arg\max} \sum_{i=1}^n
\bfw^\top \phi_\bftau(\bfx_i) y'_i\\
\end{aligned}
\end{equation}
where $\bfw\in \mathbb{R}^{d'}$ is the  parameter
vector.
Actually, it is equal to the following prediction results,

\begin{equation}
\begin{aligned}
h_\bfw(X)=sign\left ( \bfw^\top \phi_\bftau(X) \right )
\end{aligned}
\end{equation}
where $sign\left ( \cdot\right )$ is an element-wise $sign$ operation function.

To avoid the over-fitting problem, we try to reduce the complexity of the hypotheses function parameter $\bfw$ by minimizing the squared $\ell_2$ norm,

\begin{equation}
\begin{aligned}
\underset{\bfw,\xi,\bftau}{\min}~
&
\left \{
\frac{1}{2} \|\bfw\|_2^2 = \frac{1}{2}\bfw^\top \bfw \right \}
\end{aligned}
\end{equation}
We also want to reduce the prediction error of the hypotheses function on the training set. To measure the prediction error, a loss function can be applied to compare the true class label tuple $\bfy$ against the output of the hypotheses function  $h_\bfw(X)$.
The following optimization problem is obtained with a $\Delta(\bfy,h_\bfw(X))$,

\begin{equation}
\label{equ:delta1}
\begin{aligned}
\underset{\bfw}{\min}~
\Delta(\bfy,h_\bfw(X)).
\end{aligned}
\end{equation}
Instead of trying to optimize $\Delta(\bfy,h_\bfw(X))$ directly, we try to find its upper boundary and then minimize its upper boundary. Given (\ref{equ:h_w}), we have the following inequalities,

\begin{equation}
\begin{aligned}
&\bfw^\top \phi_\bftau(X)  h_\bfw(X) \geq \bfw^\top \phi_\bftau(X) \bfy', \forall \bfy'\in  \{+1,-1\}^{n}\\
&\Rightarrow
\Delta(\bfy,h_\bfw(X))+
\bfw^\top \phi_\bftau(X) \left( h_\bfw(X) - \bfy \right)\geq
\Delta(\bfy,h_\bfw(X))
\end{aligned}
\end{equation}
Thus we have the upper boundary of $\Delta(\bfy,h_\bfw(X))$, and the optimization problem in  (\ref{equ:delta1}) can be relaxed to

\begin{equation}
\label{equ:delta2}
\begin{aligned}
\underset{\bfw}{\min}~
\left \{
\Delta(\bfy,h_\bfw(X))+
\bfw^\top \phi_\bftau(X) \left( h_\bfw(X) - \bfy \right)
\right \}.
\end{aligned}
\end{equation}
We further relax the minimization  of $\Delta(\bfy,h_\bfw(X))+
\bfw^\top \phi_\bftau(X) \left( h_\bfw(X) - \bfy \right)$ to the minimization of its upper boundary, which could be obtained by exploring the class label tuple space excluding $\bfy$, $\bfy'_l\in \mathcal{Y}/\bfy$,

\begin{equation}
\begin{aligned}
\Delta(\bfy,h_\bfw(X))&+
\bfw^\top \phi_\bftau(X) \left( h_\bfw(X) - \bfy \right)\\
&\leq
\underset{l:\bfy'_l\in \mathcal{Y}/\bfy}{\max}
\left [
\Delta(\bfy,\bfy'_l)+
\bfw^\top \phi_\bftau(X) \left( \bfy'_l - \bfy \right)
\right ]
\end{aligned}
\end{equation}
Thus we can translate the problem in (\ref{equ:delta2}) to (\ref{equ:delta3}),

\begin{equation}
\label{equ:delta3}
\begin{aligned}
\underset{\bfw}{\min}~
\left \{
\underset{l:\bfy'_l\in \mathcal{Y}/\bfy}{\max}
\left [
\Delta(\bfy,\bfy'_l)+
\bfw^\top \phi_\bftau(X) \left( \bfy'_l - \bfy \right)
\right ]
\right \}.
\end{aligned}
\end{equation}
It could be further relaxed by introducing a nonnegative slack variable $\xi$ to represent the upper boundary, so that the problem could be rewritten as

\begin{equation}
\label{equ:delta4}
\begin{aligned}
\underset{\bfw,\xi}{\min}~
&
\xi,\\
s.t.~
&
\Delta(\bfy,\bfy'_l)+ \bfw^\top \phi_\bftau(X) (\bfy'_l-\bfy)
\leq
\xi,\forall l:\bfy'_l\in \mathcal{Y}/\bfy, \\
&\xi\geq 0.
\end{aligned}
\end{equation}

Combining the problems in (\ref{equ:h_w}) and (\ref{equ:delta4}), and introducing constrains on $\bftau$ to prevent negative kernel weights, the following overall optimization problem,

\begin{equation}
\begin{aligned}
\underset{\bfw,\xi,\bftau}{\min}~
&
\frac{1}{2} \bfw^\top \bfw
+ C \xi,\\
s.t.~
&
\Delta(\bfy,\bfy'_l)+ \bfw^\top \phi_\bftau(X) (\bfy'_l-\bfy)
\leq
\xi,l:\bfy'_l\in \mathcal{Y}/\bfy, \\
&\xi\geq 0,~\sum_{m=1}^M  \tau_m=1, \tau_m\geq 0, m=1,\cdots,M.
\end{aligned}
\end{equation}
where $C$ is a tradeoff parameter.

\subsection{Optimization}

To optimize this problem, we give the primal Lagrangian function as follows,

\begin{equation}
\begin{aligned}
&\mathcal{L}(\bfw,\xi,\bftau,\bfalpha,\beta,\gamma,\bfdelta)=
\frac{1}{2} \bfw^\top \bfw
+ C \xi\\
&+\sum_{l:\bfy'_l\in \mathcal{Y}/\bfy}
\alpha_l
\left (
\Delta(\bfy,\bfy'_l)+ \bfw^\top \phi_\bftau(X) (\bfy'_l-\bfy)
-
\xi
\right )\\
&-\beta\xi -\gamma\left(
\sum_{m=1}^M  \tau_m-1
\right)
-\sum_{m=1}^M \delta_m \tau_m
\end{aligned}
\end{equation}
where $\alpha_l\geq 0$, $\beta \geq 0$, $\gamma \geq 0$ and $\delta_m\geq 0$ are the Lagrange multipliers.
We argue the following dual optimization problem,

\begin{equation}
\label{equ:Lag1}
\begin{aligned}
\underset{\bfalpha,\beta,\gamma,\bfdelta}{\max}
~&
\underset{\bfw,\xi,\bftau}{\min}
~
\mathcal{L}(\bfw,\xi,\bftau,\bfalpha,\beta,\gamma,\bfdelta)\\
s.t.
~&
\alpha_l\geq 0, l:\bfy'_l\in \mathcal{Y}/\bfy,\\
& \beta \geq 0,\gamma\geq 0, \delta_m\geq 0,m=1,\cdots,M.
\end{aligned}
\end{equation}
By setting the digestives of the Lagrange function with regard to $\bfw$ and $\xi$ to zero, we have

\begin{equation}
\begin{aligned}
\frac{\partial \mathcal{L}}{\partial \bfw}=0
\Rightarrow
&\bfw=
\sum_{l:\bfy'_l\in \mathcal{Y}/\bfy} \alpha_l \phi_\bftau(X) (\bfy-\bfy_l')\\
\frac{\partial \mathcal{L}}{\partial \xi}=0
\Rightarrow
&C-\sum_{l:\bfy'_l\in \mathcal{Y}/\bfy} \alpha_l - \beta = 0
\Rightarrow
C  \geq \sum_{l:\bfy'_l\in \mathcal{Y}/\bfy} \alpha_l.
\end{aligned}
\end{equation}
By substituting these results and the kernel definition in (\ref{equ:Ker1}) to (\ref{equ:Lag1}),  we obtain the dual Lagrangian function,

\begin{equation}
\begin{aligned}
&\mathcal{P}(\bftau,\bfalpha,\gamma,\bfdelta)\\
& =
-\frac{1}{2}
\sum_{l,k:\bfy'_l,\bfy'_k\in \mathcal{Y}/\bfy}
\alpha_l \alpha_k \left (
(\bfy-\bfy'_l)^\top
\sum_{m=1}^M \tau_m^2
K_m(X,X)
(\bfy-\bfy_k') \right )\\
&
+\sum_{l:\bfy'_l\in \mathcal{Y}/\bfy}
\alpha_l
\Delta(\bfy,\bfy'_l)-\gamma\left(
\sum_{m=1}^M  \tau_m-1
\right)
-\sum_{m=1}^M \delta_m \tau_m
\end{aligned}
\end{equation}
This optimization problem is then transformed to

\begin{equation}
\label{equ:Lag2}
\begin{aligned}
\underset{\bftau}{\min}
~&
\underset{\bfalpha,\gamma,\bfdelta}{\min}
~
\mathcal{P}(\bftau,\bfalpha,\gamma,\bfdelta)
\\
s.t.~
& \alpha_l\geq 0, l:\bfy'_l\in \mathcal{Y}/\bfy,~ C  \geq \sum_{l:\bfy'_l\in \mathcal{Y}/\bfy} \alpha_l,\\
&\gamma\geq 0, \delta_m\geq 0,m=1,\cdots,M.\\
\end{aligned}
\end{equation}
To solve this problem, we adopt an alternate optimization strategy. In an iterative algorithm, $\bfalpha$ and $\bftau$ with its Lagrange multipliers $\gamma$ and $\bfdelta$  are optimized alternately.

\begin{itemize}
\item \textbf{Optimizing $\bfalpha$} By fixing $\bftau$ with its Lagrange multipliers $\gamma$ and $\bfdelta$, and only considering $\bfalpha$, the optimization problem in (\ref{equ:Lag2}) is reduced to

\begin{equation}
\label{equ:alpha1}
\begin{aligned}
\underset{\bfalpha}{\max}
&
\left (
-\frac{1}{2}
\sum_{l,k:\bfy'_l,\bfy'_k\in \mathcal{Y}/\bfy}
\alpha_l \alpha_k \left (
(\bfy-\bfy'_l)^\top K_\bftau(X,X) (\bfy-\bfy_k')
\right )
\right .
\\
&
\left .
+\sum_{l:\bfy'_l\in \mathcal{Y}/\bfy}
\alpha_l
\Delta(\bfy,\bfy'_l)
\right )
\\
s.t.~
& \sum_{l:\bfy'_l\in \mathcal{Y}/\bfy} \alpha_l \leq C, \alpha_l\geq 0, l:\bfy'_l\in \mathcal{Y}/\bfy.
\end{aligned}
\end{equation}
This problem can be solved as a quadratic programming problem.

\item \textbf{Solving $\bftau$} By fixing $\bfalpha$, and only considering $\bftau$ and its Lagrange multipliers $\gamma$ and $\bfdelta$, we have the following problem,

\begin{equation}
\label{equ:tau1}
\begin{aligned}
\underset{\bftau}{\min}
~&
\underset{\gamma,\bfdelta}{\max}
\left \{
-\frac{1}{2}
\sum_{l,k:\bfy'_l,\bfy'_k\in \mathcal{Y}/\bfy}
\left ( \alpha_l \alpha_k
(\bfy-\bfy'_l)^\top
\vphantom{\sum_{i}}
\right .\right .\\
&\left .
~~~~~~\times
\sum_{m=1}^M \tau_m^2
K_m(X,X)
(\bfy-\bfy_k') \right )
\\
&
\left . -\gamma\left(
\sum_{m=1}^M  \tau_m-1
\right)
-\sum_{m=1}^M \delta_m \tau_m
\right \}\\
s.t.~
& \gamma\geq 0, \delta_m\geq 0,m=1,\cdots,M.\\
\end{aligned}
\end{equation}
This is the dual form of a constrained quadratic programming problem, and we can solve it as a constrained quadratic programming problem.

\item \textbf{Updating $\mathcal{Y}/\bfy$}
Moreover, it should be noted that the construction of set $\mathcal{Y}/\bfy$ is also a problem.
To this end, we propose to construct $\mathcal{Y}/\bfy$ sequentially in the iterative algorithm.
We propose to construct $\mathcal{Y}/\bfy$ by adding one new class label tuple to $\mathcal{Y}/\bfy$ in each iteration according to updated $\bfw$ and $\bftau$,

\begin{equation}
\label{equ:label1}
\begin{aligned}
\bfy^*
&=\underset{\bfy''\in  \{+1,-1\}^{n}, \bfy''\neq \bfy, \bfy''\notin \mathcal{Y}/\bfy} {\arg\max}
\left \{\Delta(\bfy,\bfy'')+
\vphantom{\sum_{j:j \neq i}}
\right. \\
&\left.
\sum_{l:\bfy'_l\in \mathcal{Y}/\bfy}
\left (\alpha_l (\bfy-\bfy_l')^\top K_\bftau(X,\bfx_i) \bfy''
\right )
\right \}.
\end{aligned}
\end{equation}
where $K_\bftau(X,\bfx_i) = [K_\bftau(\bfx_1,\bfx_i), \cdots, K_\bftau(\bfx_n,\bfx_i)]^\top\in \mathbb{R}^{n \times 1}$.
Then we can update $\mathcal{Y}/\bfy$ by adding $\bfy^*$ to it,

\begin{equation}
\label{equ:Y}
\begin{aligned}
\mathcal{Y}/\bfy
\leftarrow
\{\bfy^*\} \cup \mathcal{Y}/\bfy.
\end{aligned}
\end{equation}
\end{itemize}

\subsection{Algorithm}

The iterative multi-kernel learning algorithm to optimize multivariate performance measure is summarized in Algorithm \ref{alg:MKMPM}.

\begin{algorithm}[htb!]
\caption{Multi-Kernel Learning algorithm for optimize multivariate Performance measure Optimization (MKLPO).}
\label{alg:MKMPM}
\begin{algorithmic}
\STATE \textbf{Input}: Training sample feature matrix $X$, and corresponding class label tuple $\bfy$;

\STATE Initialize $\bfalpha^0$ and $\bftau^0$;

\STATE Initialize $\mathcal{Y}/\bfy= \emptyset$;

\FOR{$t=1,\cdots,T$}

\STATE Obtain a predicted class label tuple $\bfy^*$ as in (\ref{equ:label1}) by fixing $\bfalpha^{t-1}$ and $\bftau^{t-1}$, and add it to $\mathcal{Y}/\bfy$ as in (\ref{equ:Y});

\STATE Update $\bfalpha^t$ by solving (\ref{equ:alpha1}) and fixing $\bftau^{t-1}$;

\STATE Update $\bftau^t$ by solving (\ref{equ:tau1}) and fixing $\bfalpha^t$;

\ENDFOR

\STATE \textbf{Output}:
Output the learned $\bfalpha^T$ and $\bftau^T$.
\end{algorithmic}
\end{algorithm}

\section{Experiments}
\label{sec:exp}

\subsection{Experiment I: Allergen prediction}

In the first experiment, we perform the proposed to the problem of allergen prediction to optimized various prediction performance measures \cite{dang2014allerdictor}.

\subsubsection{Dataset and protocol}

In this experiment, we used a dataset constructed by Dang and Lawrence \cite{dang2014allerdictor}.
This dataset contains 42,977 protein sequences, 3,907 of them are allergens while the remaining 39,070 are non-allergens. To extract feature from each protein sequence, we used the bag-of-words method \cite{wang2013joint}.
Firstly, the amino acid sequence of a protein is broken to some overlapping peptides with a small sliding window, and each peptides is treated as a word.
To conduct the experiment, we perform the popular 10-fold cross validation.
Various performance measures are considered in this experiment. The multivariate performance measures are optimized on the training set and tested on the test set, including AUC, RP-BEP, ACC, F score and MCC.

\subsubsection{Results}

\begin{figure}[htb!]
\centering
\subfigure[AUC]{
\label{fig:Allergen_AUC}
\includegraphics[width=0.23\textwidth]{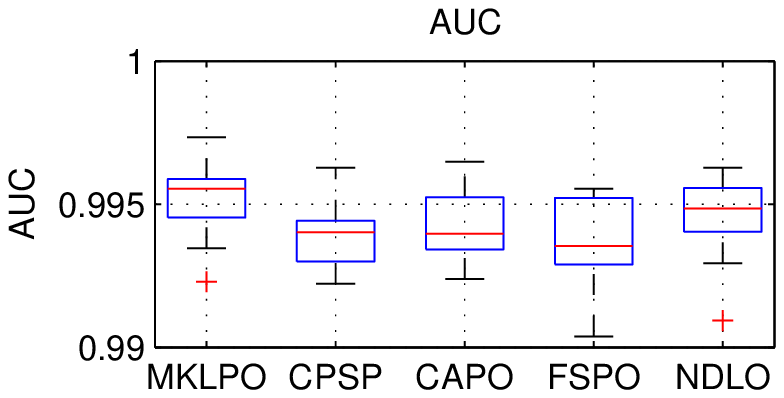}}
\subfigure[PR-BEP]{
\label{fig:Allergen_RPBEP}
\includegraphics[width=0.23\textwidth]{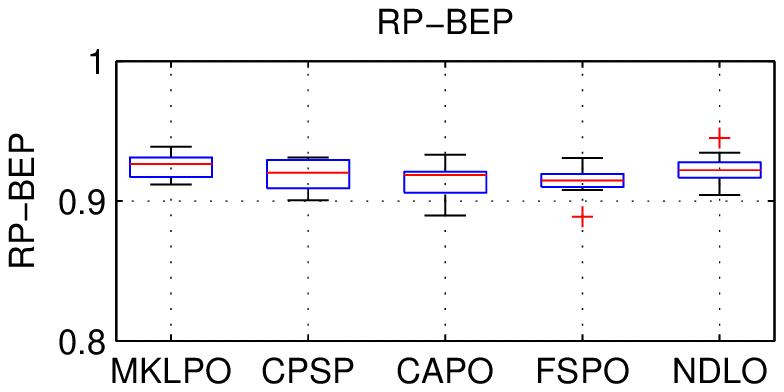}}
\subfigure[ACC]{
\includegraphics[width=0.23\textwidth]{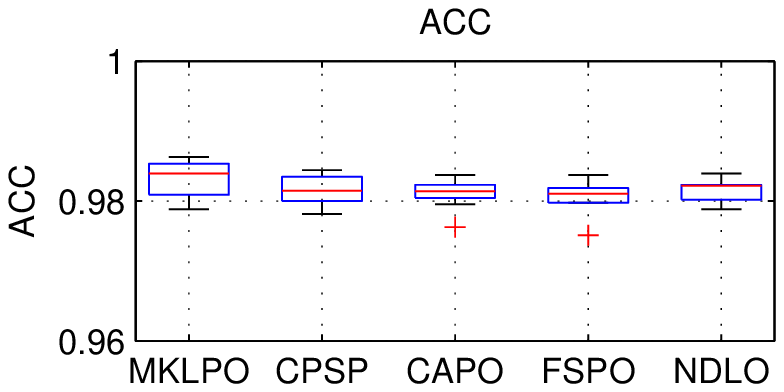}}
\subfigure[F score]{
\label{fig:Allergen_F}
\includegraphics[width=0.23\textwidth]{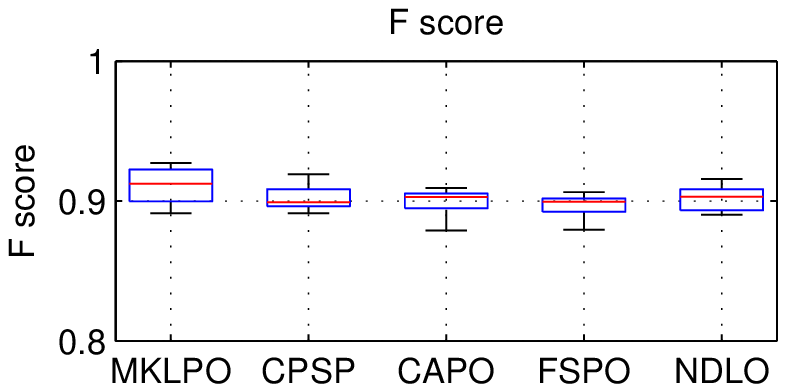}}
\subfigure[MCC]{
\label{fig:Allergen_MCC}
\includegraphics[width=0.23\textwidth]{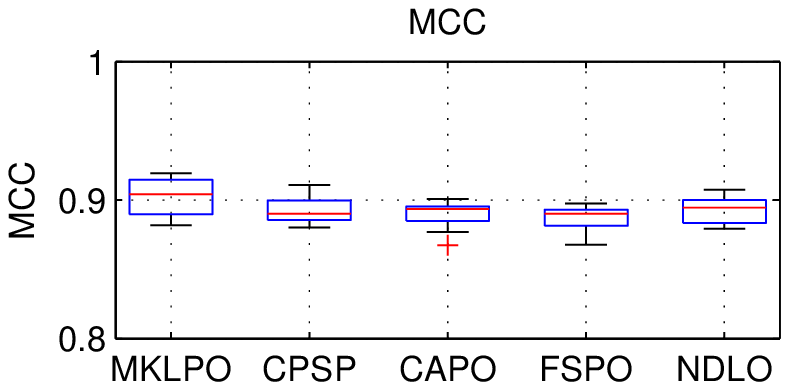}}\\
\caption{Boxplots of optimized multivariate performance measures of 10-fold cross validations of allergen prediction problem.}
\label{fig:Allergen}
\end{figure}

We compare the proposed multi-kernel learning based multivariate performance measures optimization algorithm agains the original kernel version of  SVM$^{Perf}$, cutting-plane subspace pursuit (CPSP) algorithm  \cite{Joachims2009179}. Moreover, three different variations of SVM$^{Perf}$ are also compared as the state-of-the-art multivariate performance measures optimization methods, including the performance measure optimization method by classifier adaptation (CAPO) \cite{Li20131370}, the feature selection method for multivariate performance measures optimization (FSPO) \cite{mao2013feature}, and the  non-decomposable loss functions optimization method (NDLO) \cite{ranjbar2013optimizing}.
We used these methods to optimize the multivariate performances of AUC of ROC, PR-BEP of recall-precision curve, ACC, F score, and MCC respectively on the training set, and the test them on the test set. The boxplots of the corresponding performance measures of 10-fold cross validations are given in Figure \ref{fig:Allergen}.
From this figure, we can see clearly that the proposed multi-kernel based multivariate performance measure optimization method achieves the best results with regard to different performance measures.
Similar phenomenon can be observed in Figure \ref{fig:Allergen_MCC}, and MKLPO is the only algorithm which obtains a higher MCC median value than 0.900.
For other performance measures, MKLPO also optimize them to achieve the best performances measures on the test sets.
Among the compared algorithms, both CPSP and CAPO are improved by using kernel trickles. However, due to the limitation of single kernel, their performance are not necessarily superior to the linear models, FSPO and NDLO. In most cases, their performances are comparable to each other.

\subsection{Experiment II: Rehabilitative speech treatment assessment}

In this experiment, we test the proposed algorithm for the automatic assessment of rehabilitative speech treatment.

\subsubsection{Dataset and protocol}

In this experiment, we use the dataset provided by Tsanas et al.  \cite{Tsanas20140417224724}.
There are $126$ phonations in the data set. A speech expert is employed to assess the phonations, and label them as ``acceptable" or ``unacceptable". Among the 126 phonations, 42 is labeled as ``acceptable" while the remaining 84 is labeled as ``unacceptable". Each phonation is defined as a data sample in the problem of pattern classification, and `acceptable" phonation is defined as positive sample, while ``unacceptable" phonation as negative sample.
For the purpose of pattern classification, we extract features from each of the phonations.
To conduct the experiment, we also use the 10-fold cross validation. The multivariate performance measures are optimized on the training set and tested on the test set, including AUC, RP-BEP, ACC, F score and MCC.

\subsubsection{Results}

\begin{figure}[htb!]
\centering
\subfigure[AUC]{
\label{fig:Speech_AUC}
\includegraphics[width=0.23\textwidth]{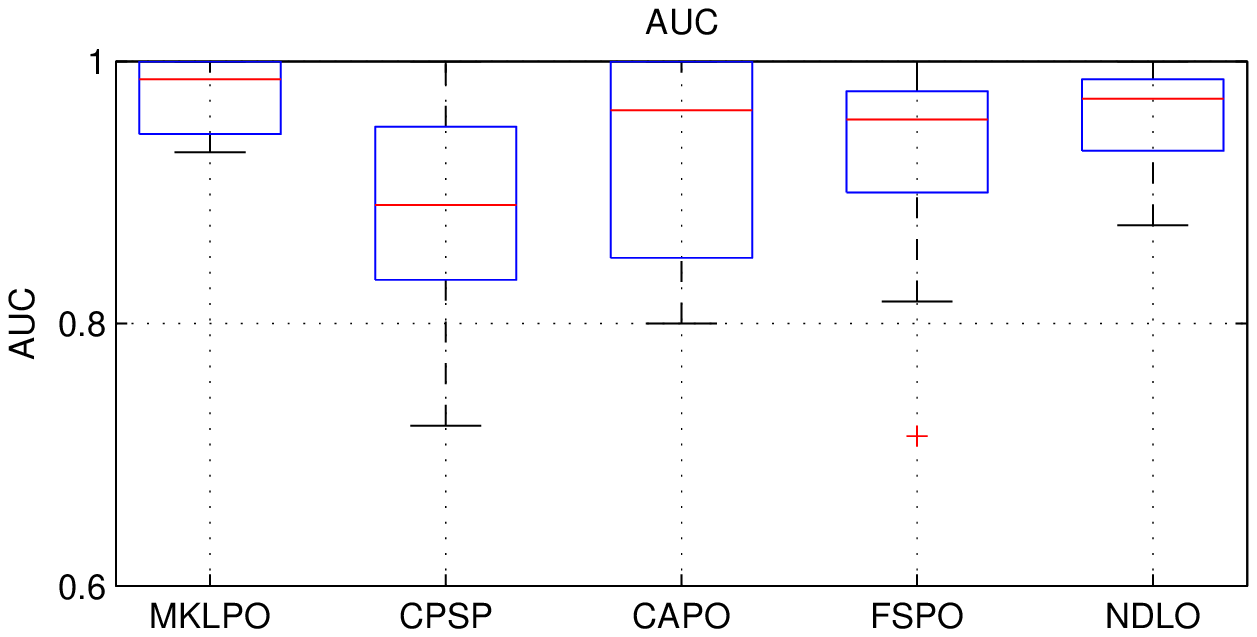}}
\subfigure[PR-BEP]{
\label{fig:Speech_RPBEP}
\includegraphics[width=0.23\textwidth]{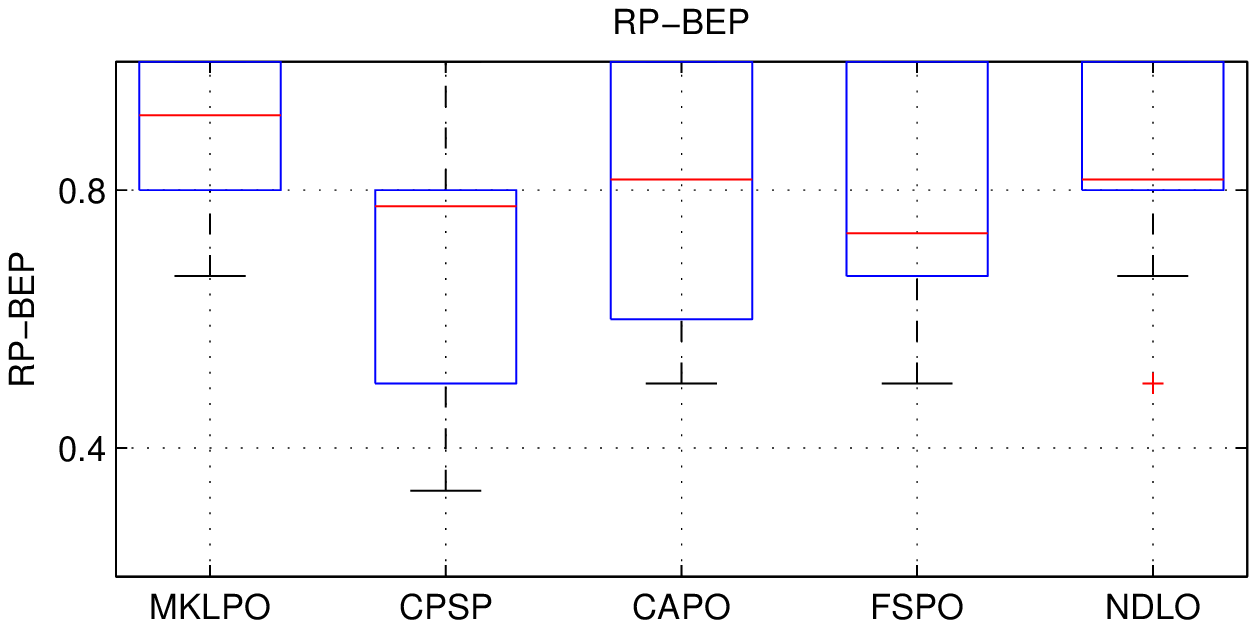}}
\subfigure[ACC]{
\includegraphics[width=0.23\textwidth]{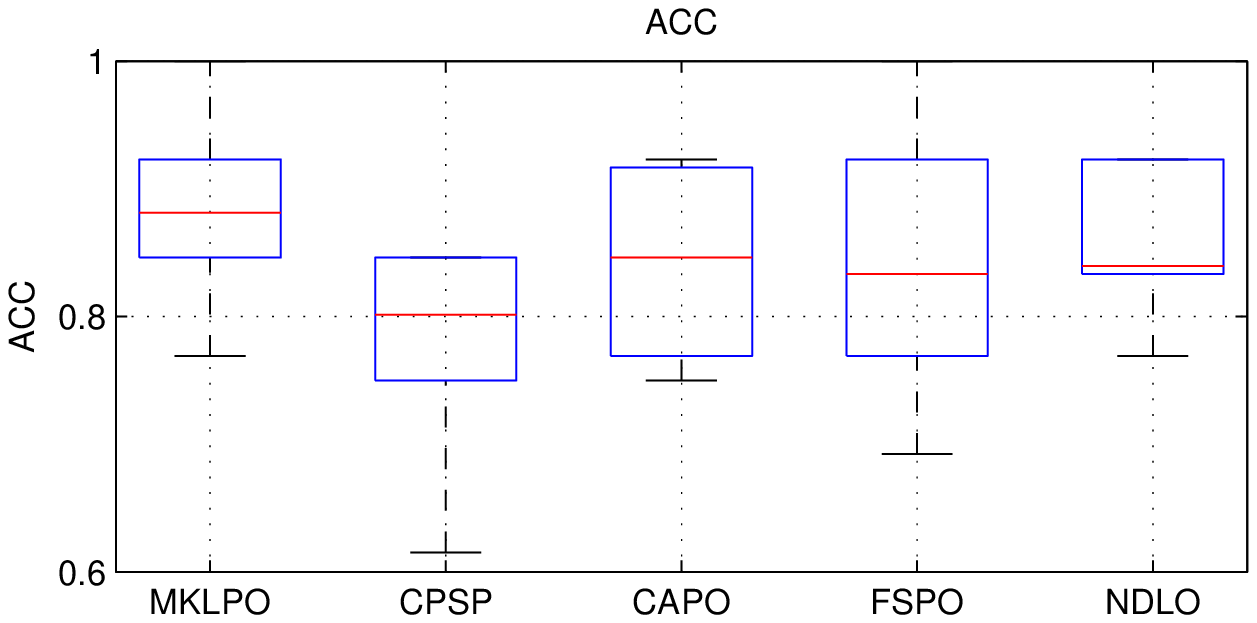}}
\subfigure[F score]{
\label{fig:Speech_F}
\includegraphics[width=0.23\textwidth]{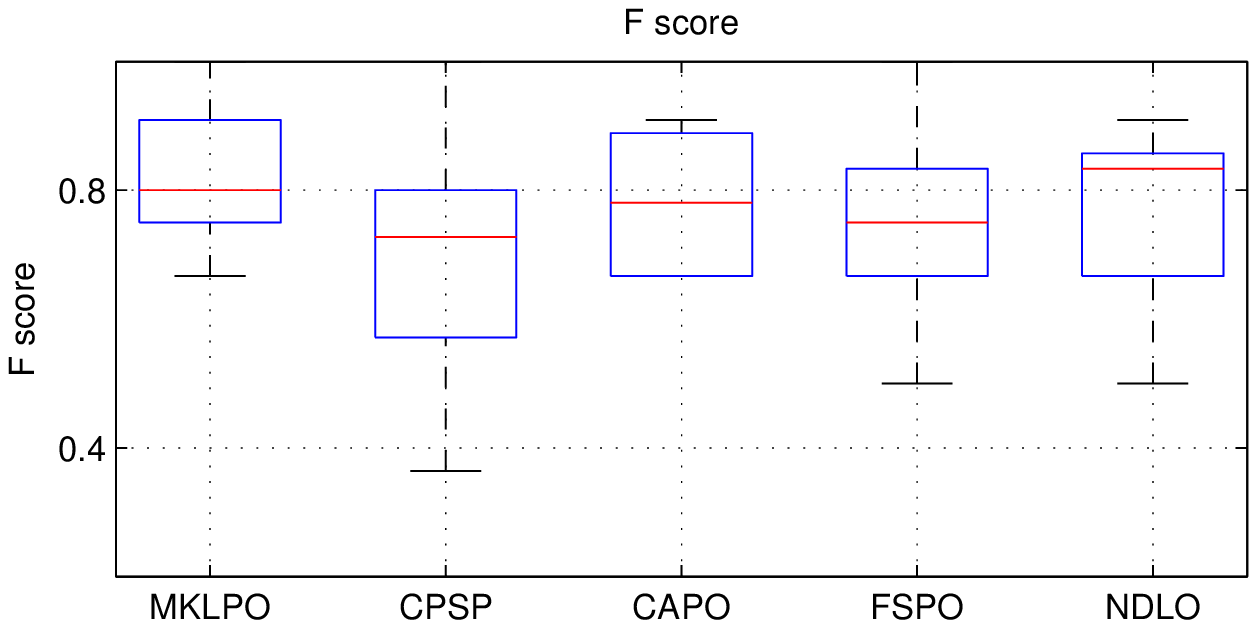}}
\subfigure[MCC]{
\label{fig:Speech_MCC}
\includegraphics[width=0.24\textwidth]{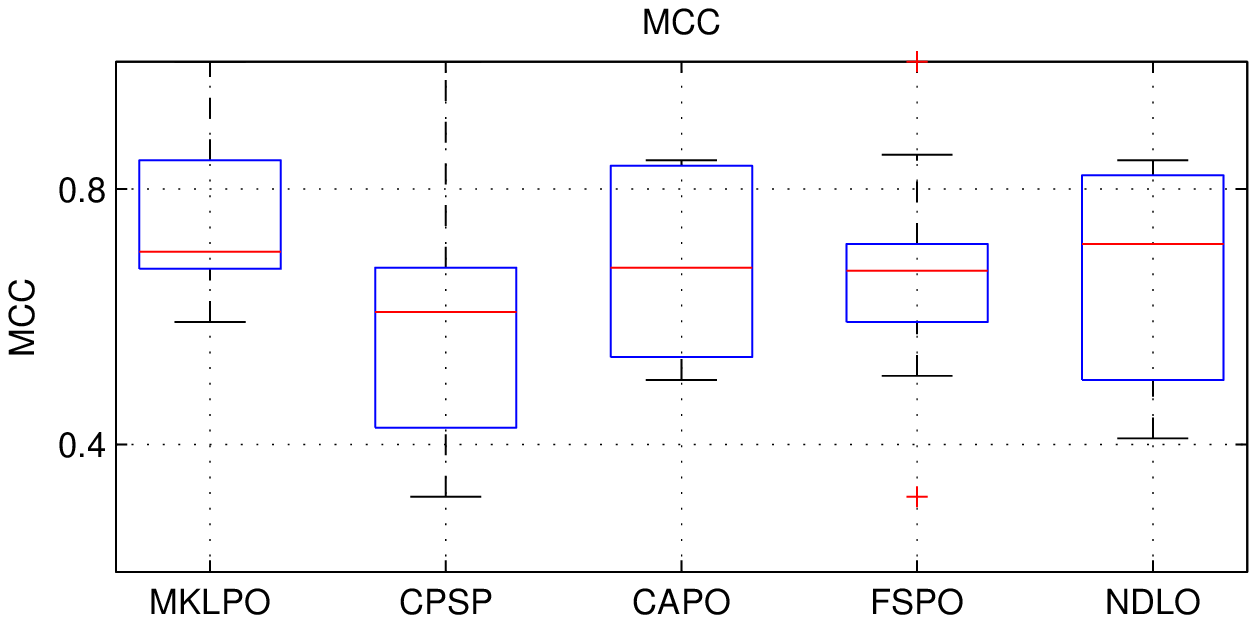}}\\
\caption{Boxplots of optimized multivariate performance measures of 10-fold cross validations of rehabilitative speech treatment assessment problem.}
\label{fig:Speech}
\end{figure}

Fig. \ref{fig:Speech} shows the boxplots of optimized multivariate performance measures of 10-fold cross validations by using rehabilitative speech treatment assessment data set.
As can be seen, our MKLPO algorithm significantly outperforms the other multivariate performance measures optimization algorithms in most cases. The performance difference is larger as the MCC is optimized as the desired multivariate performance measure. The CAPO algorithm outperforms other algorithms in most cases slightly besides the proposed MKLPO algorithm. This result is consistent with the experiment results given in the previous section.

\section{Conclusions and future works}
\label{sec:con}

Recently a multivariate performance measures optimization method is proposed to estimate a given complex multivariate performance measure as a linear function. This method is based on kernel trick. However, it is difficult to choose a suitable kernel function with its corresponding parameter. To solve this problem, in this paper, we proposed the first multi-kernel learning based algorithm for the problem of optimization of multivariate performance measures. We build a unified objective function for the learning of both multiple kernel weight and classifier parameter for the purpose of multivariate performance measure. An iterative algorithm is developed to optimize the objective function. The experiment results on two different pattern classification problems show that the proposed algorithm outperforms the state-of-the-art multivariate performance measure optimization methods. In the future, we will also explore the potential of using the proposed methods to bioinformatics problems \cite{chen2013formation,chen2009nanoparticle,wang2012proclusensem,yi2015three,wang2010conceptual,peng2015modeling,liu2013structure,zhou2014biomarker,luo2011piecewise}, integrated circuit design \cite{zhang2013eot,zhang2009new,zhang2008impacts,zhang2012germanium,zhang2008comprehensive,zhuge2009random,kang2009investigations,wang2008experimental,zhang2013accelerating,gao2014sparse}, multiple model big data analysis \cite{wang2015supervised,li2013zht,li2013distributed,wang2013using,wang2015towards,wang2014next,wang2014optimizing,zhao2014fusionfs}, software and network security \cite{sun2012unsupervised,sun2013multi,sun2014human,sun2014mobile,yi2014multispectral,zhang2011empirical,zhang2014after,zhang2011effective,huang2011distilling,zhuang2013investigating}, and power systems optimization \cite{che2014dc,che2014only}. Moreover, we will also improve the proposed method by regularizing the learning of classifier by graphs \cite{wang2012adaptive,wang2012multiplegraph,Shen20141286,shen2015transaction,wang2015nonlinear,wang2015constrained,liu2015supervised,wang2015representing,wang2015image}.

%\bibliographystyle{IEEEtran}
%\bibliography{MultiK_MultiPerf}


\end{document}